\documentclass[5p,preprint]{elsarticle}
\usepackage[numbers]{natbib}
\usepackage{graphicx}
\usepackage{subfigure}
\usepackage{amsmath,amssymb,amsfonts}
\usepackage{textcomp}
\usepackage[dvipsnames]{xcolor}
\usepackage{multirow}
\usepackage{hanging}
\usepackage{floatrow}
\usepackage{placeins}
\usepackage{subfigure}
\usepackage{lineno,hyperref}
\hypersetup{
   colorlinks=false,
    pdfborder={0 0 0},
}
\usepackage{caption}
\usepackage{changepage}
\usepackage{footnote}

\journal{MSSP}

\begin{document}

\begin{frontmatter}

\title{Oversampling Adversarial Network for
Class-Imbalanced Fault Diagnosis}

\author{Masoumeh Zareapoor \corref{mycorrespondingauthor}, Pourya Shamsolmoali , Jie Yang \corref{mycorrespondingauthor}}
\cortext[mycorrespondingauthor]{Corresponding:jieyang@sjtu.edu.cn;\\  mzarea222@gmail.com; pshams55@gmail.com}

\address{School of Electronic, Information and Electrical Engineering \\ Shanghai Jiao Tong University, Shanghai, China.}


%

\begin{abstract}
The collected data from industrial machines are often imbalanced, which poses a negative effect on learning algorithms. However, this problem becomes more challenging for a mixed type of data or while there is overlapping between classes. Class- imbalance problem requires a robust learning system which can timely predict and classify the data. We propose a new adversarial network for {\em simultaneous classification and fault detection}. In particular, we restore the balance in the imbalanced dataset by generating faulty samples from the proposed mixture of data distribution. We designed the discriminator of our model to handle the generated faulty samples to prevent outlier and overfitting. We empirically demonstrate that; {\tt (i)} the discriminator trained with a generator to generates samples from a mixture of normal and faulty data distribution which can be considered as a {\em fault detector}; {\tt (ii)}, the quality of the generated faulty samples outperforms the other synthetic resampling techniques. Experimental results show that the proposed model performs well when comparing to other fault diagnosis methods across several evaluation metrics; in particular, coalescing of generative adversarial network (GAN) and feature matching function is effective at recognizing faulty samples.  
\end{abstract}


\end{frontmatter}


\section{Introduction}
Recently, industry applications changed to electronic machines. When the manufacturing machines perform undesirable, failures will be occurred, which may lead to disastrous malfunctions such as economic losses and environmental pollution \cite{xia2017fault}. Therefore, timely diagnosis of the failures is very important to prevent extreme costs for operations and maintenance. On the other side, the amount of data generated in industrial fields is increasing day by day. This massive amount of data requires a robust learning system that can predict and classify the data efficiently. In the last few years, detecting machine failure has been a hot research topic for engineering areas \cite{zhang2019machine, liu2018artificial}. In this paper, we focus on the fault detection problem, in which, the objective is to successfully identify the abnormal or faulty operating conditions. The existing fault detection approaches have shown a fair amount of success and mostly are based on data-driven techniques \cite{lee2017application, wu2018integrated, xie2018imbalanced}. These approaches have higher requirements on available samples for model learning; i.e. for imbalanced class data, these approaches generalize poorly. The collected data from industrial environments usually suffer from masses of labeled data, since the data are naturally imbalanced, i.e. the genuine/normal data are more than faulty data in a real industrial environment. General classifiers when dealing with imbalanced data they emphasize the majority class samples. Data imbalanced problem has its applications in many real-world problems such as credit card fraud, biology and cancer detection, industrial fault diagnosis and many other aspects, even improper classification in these cases may have serious repercussions. Numerous methods and strategies have been developed to address the imbalancedness problem, in which sample handling is one of them. The sampling techniques are often easy to implement and usually, they fall under oversampling (increases number of minority class samples) or undersampling (delete some part of majority class samples) \cite{jia2018deep}. SMOTE \cite{chawla2002smote} is an advanced oversampling method which generates synthetic samples for the minority instances and many of its variations have been proposed \cite{han2005borderline}, and has proven to be a practical and effective solution to the imbalanced problem. In the last few years, deep learning techniques have emerged as a strong framework that is applied to various applications including, dimensionality reduction, motion capturing, speech and visual recognition and also more recently used to address challenges in fault diagnosis of industrial machines \cite{zhang2018adversarial, xia2017fault, janssens2016convolutional}. Deep learning models inherently have the capacity to deal with complex learning problems that are difficult to solve by traditional machine learning techniques. Generative adversarial network (GAN) introduced in \cite{goodfellow2014generative}, is a deep learning model which can generate realistic data from a random distribution, in a zero-sum game framework. Training of GAN tends to increase the error rate of the discriminative network. Since the introduction of GAN, it has been used on a number of tasks to boost classification accuracy by generating synthetic samples \cite{odena2017conditional, shamsolmoali2019g, salimans2016improved, radford2015unsupervised, shamsolmoali2020imbalanced}. 

\begin{figure*}
\hbox{\hspace{4.7em} \includegraphics[width=14cm]{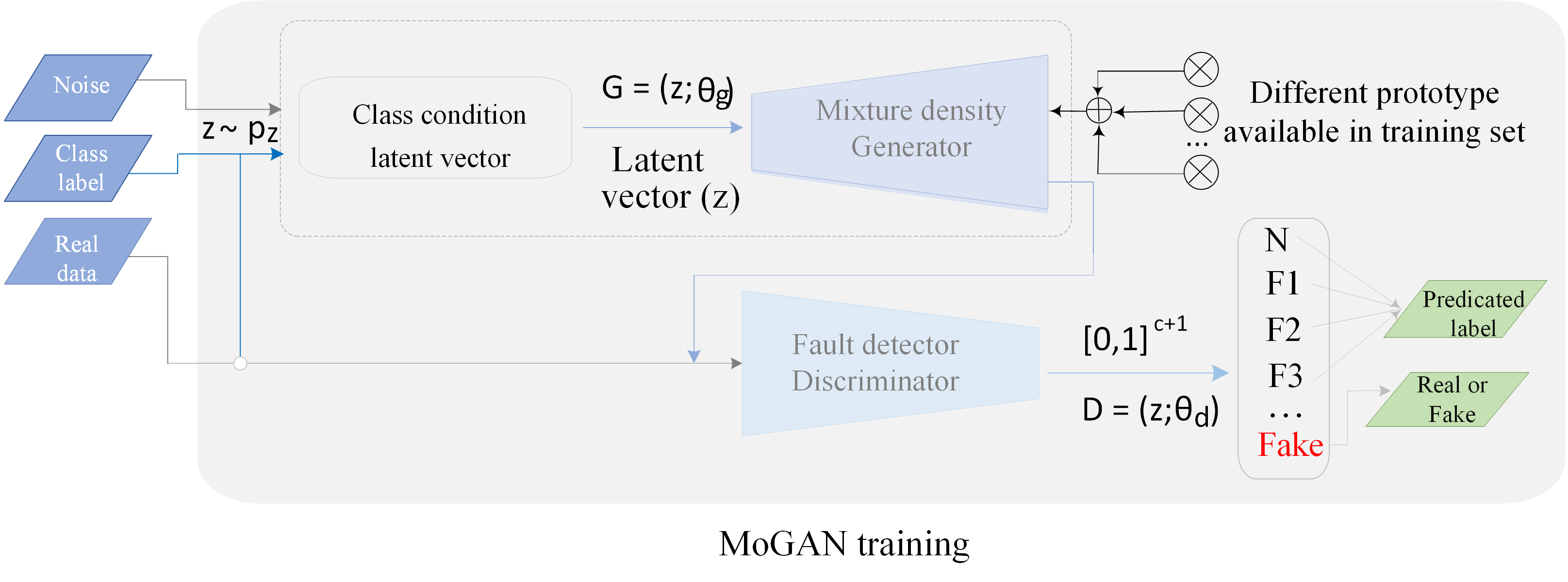}}
\caption{The {\tt MoGAN}  architecture. The discriminator of MoGAN is designed for simultaneous classification and fault detection. $e_r$ and $e_f$ indicates as real and fake (generated) samples respectively. Our discriminator creates several clusters in its embedding space to distinguish not only fake and real image but also between normal and faulty samples.}
\label{fig:1}
\end{figure*}

\begin{figure}
\includegraphics[width=8.8cm]{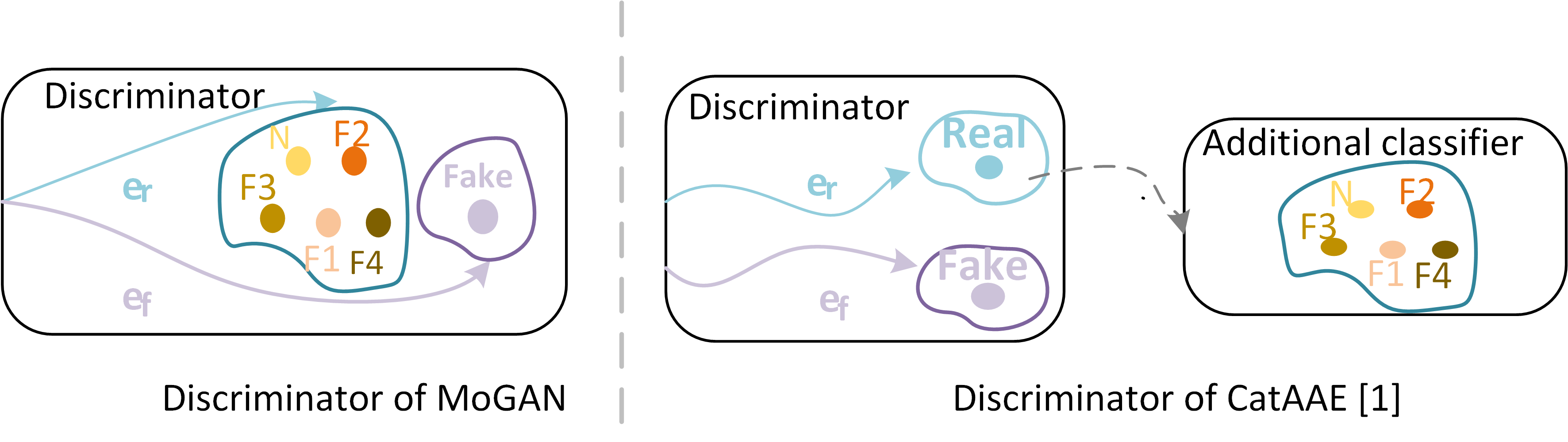}
\caption{The discriminator of our proposed MoGAN against CatAAE \cite{liu2018unsupervised}. Other methods like CatAEE in addition to discriminator; they have a classifier in their architecture to classify the faulty samples from the normal one.}
\label{fig:2}
\end{figure}

In this paper, we are interested in a model for imbalanced fault diagnosis for simultaneous classification and fault detection named {\em Minority oversampling Generative Adversarial Network} (MoGAN). The proposed model is detailed in Figure \ref{fig:1}. We used the generative adversarial network to generate synthetic and informative minority samples and successfully restore balance in our imbalanced environment. The model consists of two interdependent networks, such that, the proposed generative network is an efficient oversampling technique to generate the synthetic minority samples by incorporating majority class distribution for generating new minority samples, and the discriminative network designed to correctly classify the input into several categories. The discriminator of our model acts as a classifier and fault detector. In some variation of GAN \cite{zareapoor2019perceptual, eghbal2019mixture}, the discriminator is trained to classify the input into multiple classes rather than two classes. In this work, we embrace this idea for simultaneous classification and fault detection. During training, the discriminator learns to identify the faulty samples from the normal samples, thus, in fact, it became a fault detector (for given false positive rate). Figure \ref{fig:2} shows the structure of our discriminator, it simultaneously acts as a classifier and fault detector. The state of the arts in this regard suffers from three main problems \cite{zhang2019machine, lee2017application, razavi2017integrated, xie2018imbalanced}. First, these methods interfere with the majority of data distribution. The second is the overfitting of the training data. The third is diversity in data distribution since there are different prototype data in imbalanced datasets. 
Our proposed model conceived to address these main problems. We perform comparative results with different fault diagnosis approaches, in particular, most commonly oversampling techniques. Although the existing fault diagnosis techniques show promising results, our model improves upon these techniques across several evaluation metrics. In particular, we want to show that the discriminator of our model can be a fault detector. {\em We need to note here that, in the entire paper, we indicated the faulty samples as minority class and the normal samples as majority class}. The rest of the paper is organized as follows. Section \ref{sec:2} provides a brief background of the fault diagnosis approaches in the industrial environment. Section \ref{sec:3} and \ref{sec:4} presents the proposed fault diagnosis approach. Experiments are given in section \ref{sec:5} followed by discussion and analysis in section \ref{sec:6}, finally we conclude the model in section \ref{sec:7}.  \\

{\bf Summary}: We propose MoGAN a simulatanouse classifer and fault detector method; the generative network of our model is an efficient oversampling technique to generate the synthetic minority samples, and the discriminative network designed to correctly classify the input into several categories; the discriminator of our model acts as a classifier and fault detector. Given an imbalanced data, the extracted feature vectors are fed to the discriminator {\tt D}. D predicts the faulty label from the normal data as well as distinguishes between real and fake data
instances; {\tt N, F1, F2} and {\tt F3} indicates the normal and different type of faulty samples respectively. {\tt G} generates minority samples from the mixture
of data distribution. The class information embedded in the latent space in order to draw a target class. Note that, the proposed MoGAN while
generating a high-quality minority class sample; it also enables the discriminator to create several clusters in its output space to better distinguish
the faulty samples from the normal samples

\section{Related Works}
\label{sec:2}
In this section we discussed several approaches that are based on bearing fault diagnosis. Handling imbalancedness in fault diagnosis is one of the many ways in this field. The most recent reviews in fault diagnosis techniques can be seen in \cite{zhang2019machine, liu2018artificial}. The authors deeply investigated relevant papers in machine learning and deep learning techniques based on fault detection. 
The first work that employed CNN to fault detection approaches is proposed by Singh et al. \cite{janssens2016convolutional}. Since then many papers have appeared \cite{jia2018neural, zhang2018adversarial, zhang2019machine, chen2019mechanical, lu2017fault, jia2018deep}. These works mostly focused on a simple architecture of CNN, and the classification results significantly outperform the conventional machine learning techniques. Since our paper lies on the GAN based fault diagnosis, we mostly focused to discuss these types of methods. GAN introduced by Goodfellow et al. \cite{goodfellow2014generative} in 2014, it has achieved outstanding effects in the field of image recognition. Since then, various kinds of GAN-based algorithms have appeared. One of the earliest works that used GAN for fault detection techniques is proposed by Lee et al.\cite{lee2017application}.Their model termed as ADASYN, aims to improve the classification by addressing the class imbalance problem in fault diagnosis techniques. ADASYN method addresses the imbalanced problem by assigning a weight to the minority samples using k-nearest neighbor. CatAAE \cite{liu2018unsupervised} is another method uses GAN for unsupervised fault diagnosis approaches. The model includes an autoencoder into the adversarial training process and imposes a latent space distribution for unsupervised learning. Then it uses an additional classifier to correctly distinguish the fake distribution from the real distribution. Li et al.\cite{li2011new} proposed a fault diagnosis method to learn the most valuable features and uses these features in order to detect the abnormal behavior of the system. Feng Jia et al.\cite{jia2018neural} proposed a deep neural network technique that consists of four layers; input layer, local layer, feature layer, and output layer, to extract shift-invariant features in order to identify mechanical health fault. The authors achieved around 98\% accuracy which is higher than other existing methods. Chen et al.\cite{chen2019mechanical} proposed a new model for fault detection techniques by integrating CNN and extreme learning machine (ELM). Their proposed model consists of three steps as; pre-processing step by using continuous wavelet transforms, feature extraction step by using CNN, classification step by using ELM. More recently, a two-stage learning method contains sparse filtering and a neural network is proposed for fault diagnosis techniques in order to learn the discriminative features from raw signals \cite{lei2016intelligent}. 
More recently, Jiao et al. \cite{jiao2020residual} proposed RJAAN model for fault diagnosis when there are discrepany distributions between the source domain and the target domain. The authors employed the joint distribution alignment into their learning structure to develop a comprehensive domain adaptation framework for fault diagnosis models. In \cite{wang2020multi}, Wang et al. proposed Multi-scale transfer learning for bearing fault diagnosis model termed as "MDIAN". This model aligns the feature distributions through multi-scale strategy. This authors demonstrated theor model capacities through an extensive evaluations.  

Although these methods have shown successful performance on induction motor fault diagnosis, most of these approaches have a difficult training process and has not been investigated fully advantage of deep learning models. Some sampling-based methods are used for fault diagnosis in order to improve the classification though rebalancing the dataset. Xie et al.\cite{xie2018imbalanced} improved classification performance by generating some potential synthetic minority samples. EMICIL is proposed in \cite{razavi2017integrated} for class imbalanced learning based fault diagnosis approaches. However, these methods can loosely be considered to be similar to the proposed MoGAN, as they are not simulations techniques for classification and fault detection. 

\section{GANs and cGANs }
\label{sec:3}
The original GAN consists of two networks, generative and discriminative which are simultaneously trained. The generator trains to generate fake samples which are very similar to real samples, and the discriminator trains to distinguish the real samples from the fake samples. Given a set of sample $z$ from the real data distribution $ D_{real}=(x_i) _i^n$; let $G_u$ denotes the generators, where is often neural network in practice, and $u$ is parameters of the generators; similarly, $D_v$ denotes the discriminators, and $v$ is the parameters of the discriminator. GAN trains to obtain $\theta^{(G)}$ to generate samples from the data distribution $D_g$, and the discriminator learns to recognize whether the samples are generated $D_g$ or real data $D_{real}$. The basic GAN \cite{goodfellow2014generative} is the training parameters $u$, $v$ so as to optimize the following objective function:
\begin{equation}
\begin{split}
 \min_{G_u}\max_{D_v}E_{x\sim D_{real}} \log D(x; \theta_d)    \\
+ E_{z\sim D_z} \log(1-D(G(z; \theta_g); \theta_d))     \\ 
\label{eq:e1}
\end{split}
\end{equation}\par
where, $D_g$ and $D_z$ are the empirical distributions of training samples. For a random sample $x$, which either belong to $D_{real}$ or $D_g$ and also the parameter $\theta_d$, we have a binary class as $D(x,\theta_d )\in[0,1]$ which is a score based on the probability of x (belong to real data or generated data). In Eq.\ref{eq:e1}, the discriminator should give a high score for real samples, while minimizing it for generated samples from $D_g$. In the other side, the generator works opposite of the discriminator such that it learns to maximize the score for the fake samples; i.e., it aims to minimize the $E_{z\sim D_z }\log(1-D(G(z;\theta_g);\theta_d))$ while maximizing the $E_{z\sim P_z }\log(D(G(z;\theta_g);\theta_d))$. The c-GAN \cite{odena2017conditional} is an extension model, which contains extra information from the data and termed as the conditional generative adversarial network. The c-GAN framework can be formulated as follows. For generator G;  $G:Z\times Y\rightarrow X$ , and similarly for the discriminator  the adversarial algorithm in Eq.\ref{eq:e1} can be reformulated as: 
\begin{equation}
\begin{split}
 \min_{G}\max_{D}V(D,G)=N_D + N_G    \\
 f_D = E_(x,y~p_{real}(x,y))[\log D(x,y)]       \\ 
f_G = E_{z,p_z,y\sim {p_y}}[\log(1-D(G(z,y),y))]  \\        
\label{eq:e2}
\end{split}
\end{equation}\par
where $(x,y)\in X\times Y$ are sampled from the real data distribution $p_{real}(x,y)$, and $z\in Z$ values are from the noise distribution $p_z (z)$ and $y\in Y$ are sampled from conditional data vectors which in this paper we define $y$ as a class of training and we want to generate samples towards a specific target. In this paper, since we are dealing with different class conditioning, we embrace the "c-GAN" \cite{odena2017conditional} idea and apply the class information in the last space to draw desire samples towards a target class. We discuss the model structure in details, in the next section. 

\section{Fault Diagnoses with MoGAN}
\label{sec:4}
Inspired by the prior research, in this paper we present a new scheme for imbalanced fault diagnosis approaches. The proposed model aims to restore the balance by generating realistic minority samples from a mixture of data distribution. Resampling techniques are widely used in fault diagnosis techniques; however, they may generate unnecessary or improper samples which lead to complicating the computational cost, and also, degraded the classification results, for more details we refer the reader to \cite{zhang2019machine, liu2018artificial}. {\em We are interested in a model for simultaneous classification and fault detection}, which can be addressed by our proposed MoGAN. In this paper we show an effective way to; generate the most informative minority samples; adopt a new data distribution (a mixture of all available distribution) to avoid generating improper samples; enable the model to generate samples with a high variety; correctly classify the faulty and normal samples. In fact, we would like to propose a fault diagnosis technique that not only correctly produces samples for a target class, but also classifies the outputs as one of the classes that are available during training (either normal or any type of faulty samples). We design the discriminator in addition to its ability, to be able to return the faulty samples to its corresponding class and became a fraud detector. The architecture of the proposed model is given in Figure \ref{fig:1}. 
As we briefly explained the GAN in section \ref{sec:3}, it contains two networks; $G-network$ and $D-network$ that are trained simultaneously. However, generating samples of imbalanced data from a latent vector $z$ is difficult, and this requires an accurate estimation of probability distribution between the class samples to prevent the overlapping problem. To overcome this problem, we propose a mixture probability distribution for learning a joint distribution of both the majority and minority class samples in the adversarial training. We assume $G(x)$ is the probability of the generated sample $x$, and $D(x)$ is the probability of classifying x either as fake samples or one of the $c$ classes. However, in some formulation of GAN, the discriminator considered as a multiclass classifier \cite{kliger2018novelty, eghbal2019mixture} rather than $bi-class$. It means that, instead of having two outputs as real or fake samples, it has multiple outputs as $k+1$ and known as a {\em multiclass discriminator for simultaneous classification and fault detection}. If real training set given by $y_k\in R^R$ (where $k=0,1,2,...$) with the $c$ class, the generator G aims to generate the fake samples according to the following \cite{kliger2018novelty}:  
\begin{equation}
\tilde y_k = G(\tilde z_k, \tilde a_k; u) = u\tilde z_k + \tilde a_k
\label{eq:e3}
\end{equation}

where, $\tilde y$ indicates fake samples; $\tilde z$ is a random variable; $\tilde a$ is a distribution on different classes, and vector $u$ is the parameters of the generator. Similarly, we define the discriminator as $D(x; w) = f(x^T w)$, where $x$ is an input vector contains real and fake samples; vector $w$ represents the discriminator parameters. Let $x$ be the real sample from $p_{real}$ distribution, and $\hat x_-$, $\hat x_+$ are the majority-minority classes respectively from the marginal distribution of $p_x$. G is the generative model of our GAN, which maps a random vector $z$ to images that have the same support as $x$. We denote the distribution of $g(z)$ by $p_g(x)$. We assume that our imbalanced dataset deals with two sets class component and simply divide our true distribution into two components as; Majority and Minority data distribution. One way to find the probability distribution in an imbalanced dataset (including normal and faulty samples) is to estimate a set of the majority density, then declare test samples outside of the estimated level set as fault \cite{kliger2018novelty}. Unfortunately, these methods are not accurate since they mostly related to the one-class density (prone to majority class) and neglect estimating the minority density level, due to severe lack of the samples. However, our idea is to learn these components jointly rather than single one in order to approximate the data distribution for the minority samples. We consider the density of the data points of random class $c$ as $p_c(f)$ with a threshold $\delta_c \in[0, 1]^{(c+1)}$, and $f_c$ is the subset of data support, where $f_c={f : p_c (k) > \delta_c}$  ; such that the $f_c$  disjoint with the boundary margin if $\{\delta_c\}_{c=1}^c$. 

We assume that there are two set class components as normal (majority) $F_{Mj}$ and faulty (minority) $F_{Mi}$  class samples, $f_j \in F_{Mj}$ and $f_i \in F_{Mi}$. If we have access to both densities (normal data and faulty data) then according to \cite{kliger2018novelty, salimans2016improved} the detection rate can be achieved with a threshold ratio :
\begin{equation}
\frac{F_{Mj} (x)}{F_{Mi} (x)}
\end{equation}
In practice, for an imbalanced dataset, it is difficult to have access to the minority examples distribution, because there are not sufficient training samples for such class. Our generative model generates samples $F_{Mi}$ from a mixture distribution. Using mixture distribution for minority class samples will be in the form of
\begin{equation}
\pi f_j+(1-\pi)f_c; where \pi\in [0, 1]
\end{equation}

 This observation implies that function $f$, can be as generator $G$, then the discriminator $D$ will be optimized by the theories in the next section. 

\subsection{Theoretical analysis}
Many data sampling techniques have had a fair amount of success in recent years \cite{zhang2019machine, chen2019mechanical, liu2018artificial}. We found that, it is not easy to assist conventional GANs in imbalanced data. This is due to the fact that, with fewer minority samples, the random parameters of the generative model might generate similar samples which may lead to an overfitting problem. To overcome this problem, proposed \verb|MoGAN| uses all the available data distribution to produces the realistic samples for the minority component in the training set. To implement the proposed model, let $D$ and $G$ denote the discriminator and generator, with their corresponding distribution $P_D, P_G$. The discriminator $D$ learns an accurate decision boundary for all training set; (1) for any $(x,y)\in R$  there is $w_y^T f(x)>w_c^T f(x)$; (2) for any $x\in G$,there is $\max_{c=1}^c w_c^T f(x)<0$. In this assumption, both conditions show the classification correctness and true-fake correctness on the real data and the generated data respectively. As we mentioned above, $P_D$ is trained over $c+1$ classes, where $c$ classes are from true data and $(+1)$ is the fake detected samples. In more details, the score at $(c+1)^{th}$ index $D_{(c+1)}(.)$ indicates the probability of $p_g$ distribution, hence represent a fake sample. Moreover, as shown in \cite{dai2017good}, if a generator produces samples which meet $p_g (x)=p_r (data)$, the discriminator will be failed to identify the sample classes. However, we believe that in the case of diverse class data with an imbalancedness environment, a good generator has a form of $p_g (x)\neq p_r (data)$, which is proved in \cite{dai2017good}. In this assumption, the generator is able to generate samples by using all the available data distribution. As we mentioned above, our generative model contains mixture of data distribution 
\begin{equation}
p_g (x) = \pi p_{mj} (x)+\pi(1-p_r (x))
\end{equation}
 including majority class distribution $p_{mj}$ and minority class distribution $(1-p_r)$, suppose $Supp(p)=\cup_{c=1}^c Supp(p_g)\cup Supp(p_r)$.

\begin{equation}
\delta_j^T f_g=\pi \delta_j^T f_j+\pi(1-\delta_c^T f_c)\leq0
\label{eq:e4}
\end{equation}
Moreover, according to the above conditions, the first statement is $\delta_j^T f_j>0$ , therefore, in Eq.\ref{eq:e5}, $\delta_j^T f_g<\delta_c^T f_c$ for any $i\neq j$. Further, we estimate the probability distance as, $T_c=T(\mu_c, \Sigma c)$, where it contains mean vector $\mu_c$ and covariance matrix $\Sigma c$. this equation should stand for each class, and we compute $\mu_c$ and $\Sigma c$ for all classes and then drawn a latent vector $z_c$  from distribution $T_c$. Based on Eq. \ref{eq:e2}, the generator is trained by minimizing:
\begin{equation}
 \min_{\theta_g} E_{z\sim p_z}[\log(1-D(G(z; \theta_g); \theta_d))]   
\label{eq:e5}
\end{equation}\par
therefore, the optimal discriminator for false positive FP will be derived in the form of 
\begin{equation}
\acute D=\frac {p_r (x)}{p_r (x)+p_g (x)}
\end{equation}

where $p_r (x)$ is a true/real data distribution. In this paper we use $"r"$ to denote the real data, $"Mj"$ and $"Mi"$ also denote the normal and fault samples with their corresponding distribution as $p_{Mj}, p_{Mi}$. If we have access to $p_{Mj}$  and $p_r$, then the false positive rate will be in the form of:
\begin{equation}
\frac{p_g (x)}{p_r (x)}
\end{equation}
 Now, we state and prove some important theory in what we follow. \\

\noindent {\bf Theory 1:}  For a fixed generator G, with a mixture data distribution, the optimal discriminator $\acute D$ will be in the form of  $\acute D=\frac{p_{Mj}+(1-\pi)p_r}{P_r}$.   \\

\noindent {\bf Proof.} Given G and $V(D,G)$ can be formulated as: 
\begin{equation}
\begin{split}
\frac{1-\acute D}{\acute D} = \frac{p_g (x)}{p_r (x)}   \\
 = \frac{\pi p_{Mj}(x) + (1-\pi)p_r(x)}{p_r (x)}= \frac{\pi p_{Mj}(x)}{\pi p_r(x)}+(1-\pi)    \\ 
\label{eq:e6}
\end{split}
\end{equation}\par
then according to Eq. \ref{eq:e6}, the optimal discriminator is also a false positive rate for fault detection. It is worth to mention that, the proposed discriminator is a neural network with an output of $d-dimensional$ component (as, c-class, fake). It comprises two steps. First, discrimination between real and fake is needed, such that, the discriminator encodes the input image $x$, into $[d=\Sigma_{(i=1)}^m w_i  \times \varphi(\mu_i;\Sigma_i)]$; where $\varphi(.)$ is the Gaussian probability density function \cite{eghbal2019mixture}, $w_i$ iis the mixture weight either can be [0, 1], if the component produces the highest similarity amidst the other components, indicate value 1, or value 0 otherwise. $\Sigma_i$ indicated as covariance matrix for Gaussian component $i$. Calculating $d$, is the probability of being a real image. Furthermore, in the second step, the model assigns the corresponding class as, normal sample or faulty sample. 
\begin{equation}
\begin{split}
V(D)=[E_{(Mj\sim p_r)}\log(D(Mj ))] \\
+ [E_{(Mi\sim p_r)} \log(D(Mi ))]+ [E_{z \sim p_z)} \log(1 - D(G(z)))]   \\
= \int_x [p_{Mj}\log (d(D(x))) + p_{Mi}\log(1 - d(D(x)))] \\
\label{eq:e7}
\end{split}
\end{equation} 

\noindent {\bf Theory 2} : With $\acute D$, the optimal $G$ is attained when $Mj \sim P_{Mj}$  and $Mi \sim P_{Mi}$  have $zero$ mutual information; i.e., they should not share similar information.  \\
\begin{figure}
\includegraphics[width=9.3cm]{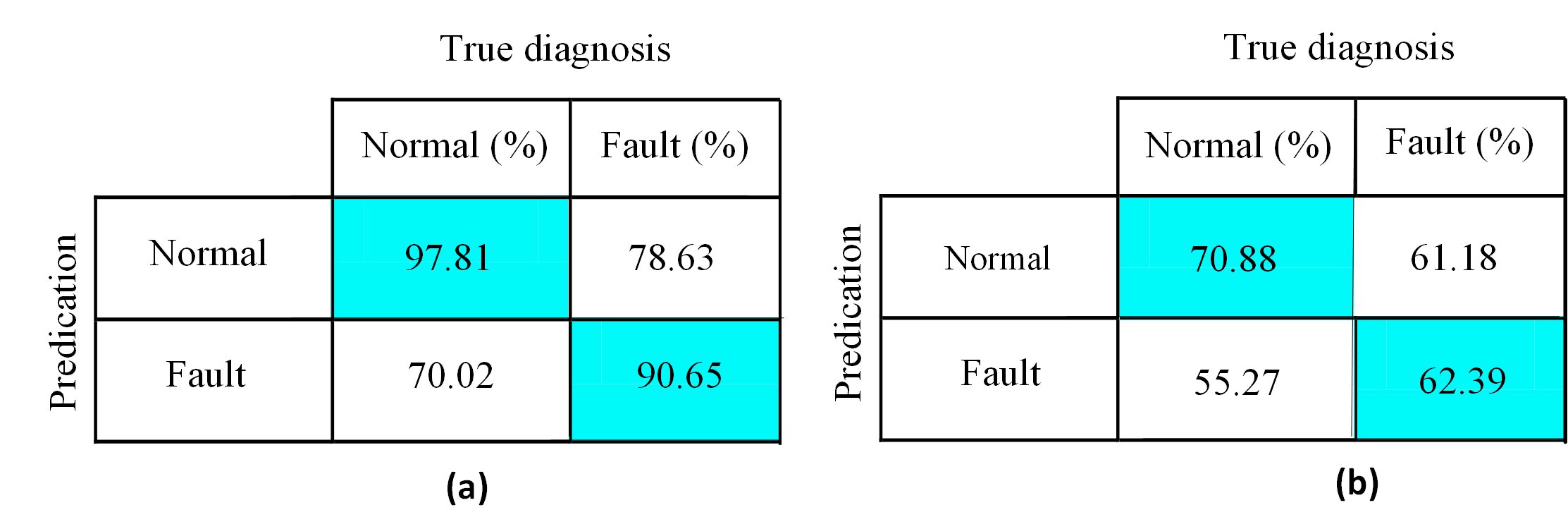}
\caption{Confusion matrix for creating synthetic dataset;{\bf  (a)} using proposed mixture data distribution and {\bf (b)} using SVM
classifier on generated data to show the robustness of the balanced dataset.}
\label{fig:3}
\end{figure}

\noindent {\bf Proof.}  This theory can be possible if the loss is $zero$. According to Eq \ref{eq:e5}, to train such mixture data distribution that could generate minority samples in low-density regions of the real data manifold, we need a loss function that efficiently encourages generator to do that. We followed \cite{dai2017good} that, the authors proposed a loss function based on KLD (Kullback-Leibler divergence) as follow:
\begin{equation}
\begin{split}
 \min_{\theta_g}-\Gamma (p_g(x; \theta_g)+E_{x\sim p_g (x; \theta_g)}\log_{P_{Mj}}(x)\\
\rho [P_{Mj}(x)>\delta]+Loss (G)   \\
where;  Loss (G) = \min_{\theta_g}\big\|E_{x\sim p_r (x)}[f(G(z; \theta_g))]\big\|     
\label{eq:e8}
\end{split}
\end{equation}\par
where $\Gamma(.) , \rho(.)$ are the marginal and joint entropy function. 

\begin{table*}[h]
\centering
\caption{The details of CWRU dataset}
\begin{tabular}{p{.08\textwidth}p{.05\textwidth}p{.05\textwidth}p{.05\textwidth}|p{.08\textwidth}p{.05\textwidth}p{.05\textwidth}p{.05\textwidth}|p{.08\textwidth}p{.05\textwidth}p{.05\textwidth}p{.05\textwidth}}
\hline 
\multicolumn{4}{c}{Data A} & \multicolumn{4}{c}{Data B} & \multicolumn{4}{c}{Data C} \\   \cline{1-12}
\hline \hline
Normal &  IR  & OR  & BF & Normal & IR & OR & BF& Normal &  IR & OR & BF  \\ [-1ex]
0.0000 & 0.007 & 0.014 & 0.021 & 0.0000 & 0.007 & 0.014 & 0.021 & 0.0000 & 0.007 & 0.014 & 0.021 \\ 
\hline 
800 & 800 & 800 & 800 & 800 & 800 & 800 & 800 & 800 & 800 & 800 & 800 \\
\hline
\end{tabular}
\label{tab:1}
\end{table*} 
\begin{table*}[h]
\centering
\caption{The details of IPF dataset}
\begin{tabular}{p{.18\textwidth}p{.09\textwidth}p{.05\textwidth}p{.05\textwidth}p{.08\textwidth}p{.05\textwidth}p{.05\textwidth}p{.05\textwidth}}
\hline
labels & Normal & F1 & F2 & F3 & F4 & F5 & F6 \\
\hline \hline 
No. of samples &  65947  & 5140  & 3381 & 3412 & 68 & 127 & 28109  \\  [0.5ex]
\hline
\end{tabular}
\label{tab:2}
\end{table*} 
\begin{figure}
    \centering
    \subfigure[]{\includegraphics[width=0.85\textwidth]{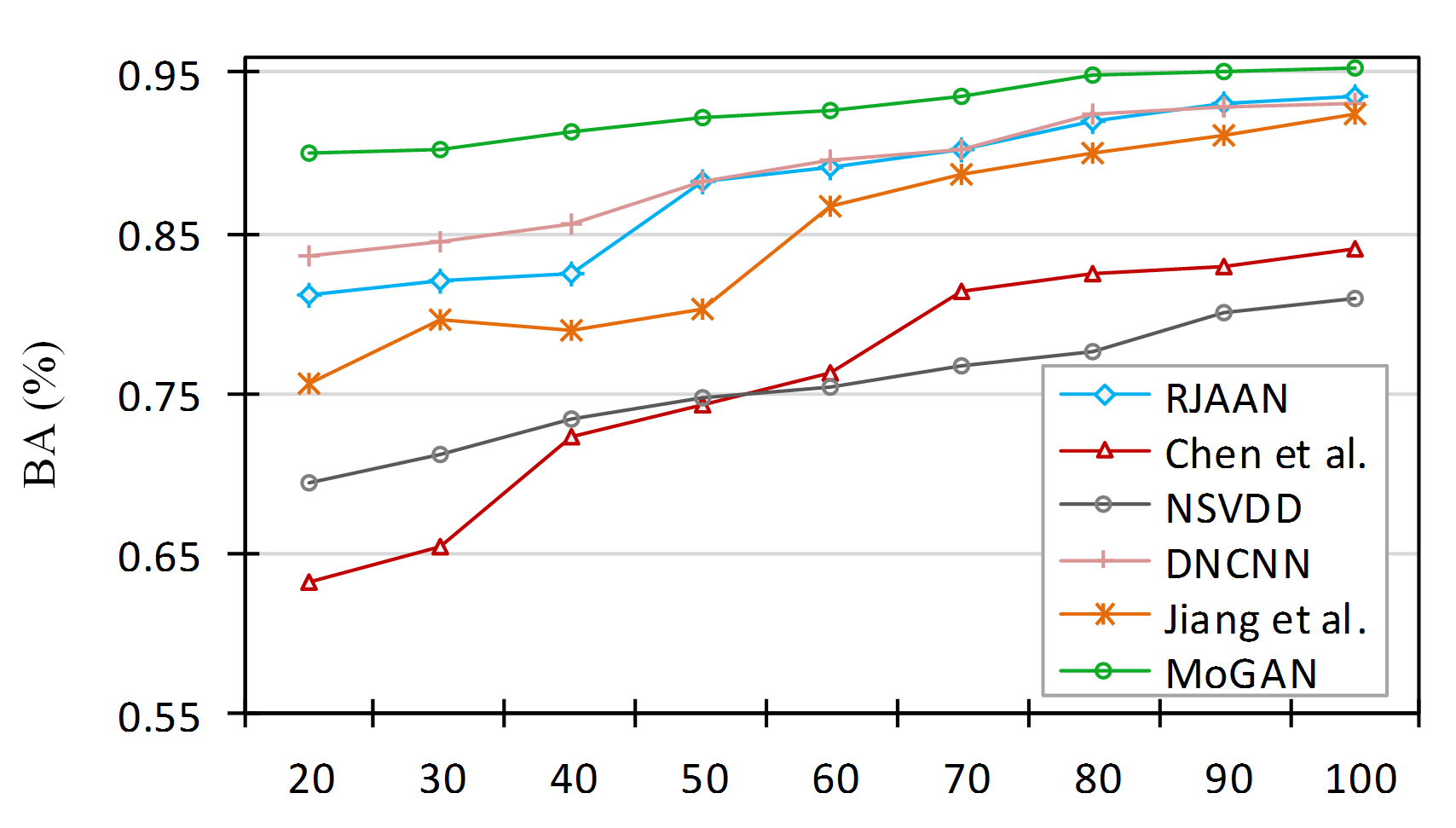}}   
    \subfigure[]{\includegraphics[width=0.85\textwidth]{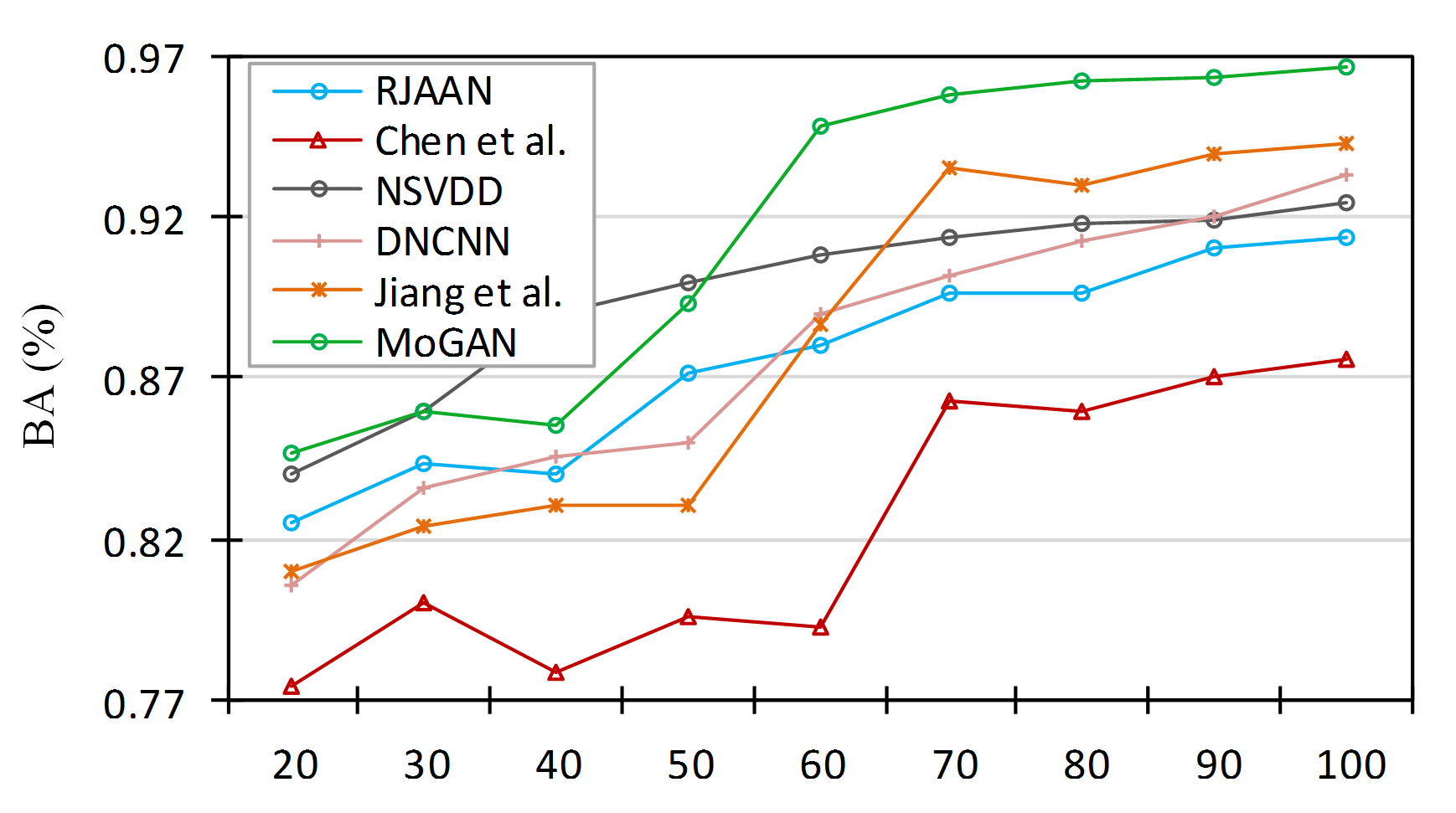}} 
       \caption{Classification Balanced accuracy (\%) of minority class with different number of hidden nodes on (a) S1 and  (b) S2 datasets. We used NSVDD \cite{liu2020semi} and DNCNN \cite{jia2018deep}, Chen et al.\cite{chen2020deep}, Jiang et al.\cite{jiang2019novel} and RJAAN \cite{jiao2020residual} as baselines.}
    \label{fig:4}
\end{figure}
It comprises three terms; produce a generator with a distribution which not intersects with a high density of real data (first \& second term), meanwhile, it is close to data manifold (third term). According to their results, the generator that is trained with Feature matching loss (Eq. \ref{eq:e5}), is able to generate multivariate samples which, falls on data manifold, and also samples which are scattered outside of data manifold. It means that when a generator is trained by feature matching loss is able to generate mixture samples from a mixture of data distribution. Theory 2, by adjusting feature matching loss, signifies that, $Mj \sim P_{Mj}$  and $Mi\sim P_{Mi}$  are completely uncorrelated. Therefore, that is what we intend; generating faulty samples (minority) from a mixture distribution which are different and recognizable from the normal samples. \\
To summarize, we proposed a new adversarial network for imbalanced fault diagnosis, to generate minority samples in a low-density area of the true data by using feature matching loss that could enrich the generator distribution and accordingly improves the detection rate of the discriminator. The main differences between this model and other GAN variations are; the way to generate the minority samples, stability, prevention of any changes in majority class, and also, prove that the discriminator can be fault detector in the case of fault diagnosis. In Section \ref{sec:5} we empirically demonstrate that when our designed discriminator trained with a mixture generator based Feature Matching loss, has an impressive ability to detect real fault samples. 

\subsection{Learning Proposed Fault Diagnosis Model}
Based on the above observations, we proposed a fault diagnosis approach which not only extracts the high level and discriminative features from raw data, but it can rebalance the training data in order to improve the model performance. Several authors applied GANs for imbalanced data classification based fault diagnosis techniques \cite{liu2018unsupervised,zhang2018adversarial, lee2017application}. Nonetheless, these approaches are not directly applied to specific classes. Our proposed MoGAN learns all the available data distribution and generates the minority class samples for the training set. Then, the obtained results from the mixture generator fed to the discriminator for the further process. For designing the generator and discriminator networks we follow the convention as DCGAN \cite{radford2015unsupervised}. However the difference is in defining the extra condition in G-network in order to draw a specific target, and also the objective function of the generative model, which we use Feature Matching loss \cite{salimans2016improved}. D-network also which is the false-positive rate for the fault diagnosis; also, it is designed to identify the wrong samples to the generator based on the corresponding classes. 

\subsection{Model training}
First, we extract all the fault and normal features as rare and normal cases. We use mixture data distribution \cite{kliger2018novelty} to generate the faulty samples and then joint the generated samples to the original corresponding samples in for balanced data. The generator and discriminator involve a $2D  transpose$ convolutional network. Both networks are
inspired by the pioneering work DCGAN \cite{radford2015unsupervised}. The generative model contains $5$ convolution transpose layers (deconvolutional layer) to perform up-sampling on a $128-dimensional$ vector from a proposed mixture distribution. We used instance normalization \cite{ulyanov2016instance} instead of using batch normalization between each of the deconvolution layers. This allows the model to increase training stability. Parametric Rectified linear units (PReLU) are used in all layers except the final layer. We replaced the tanh nonlinearity function with Feature Matching loss to the final output of the generative model. We implement our proposed model by using Keras 2.1.2, the deep learning open-source library, and TensorFlow 1.3.0 GPU as the backend deep learning engine. Python 3.6 is used for all the implementations. All the implementations of the network are conducted on a workstation equipped with an Intel i7-6850K CPU, a 64 GB Ram, and an NVIDIA GTX Geforce 1080 Ti GPU and the operating system is Ubuntu 16.04. 

\section{Experiments and Protocol}
\label{sec:5}
We perform an extensive experimental on the proposed {\tt MoGAN} and compare it to several fault diagnosis approaches. We used different oversampling techniques EWMOTE \cite{lin2018adaptive}, EMICIL \cite{razavi2017integrated}, WSMOTE \cite{zhang2016transfer} and Border-SMOTE \cite{han2005borderline} on classification faulty samples. In addition, we compare our model to A2CNN \cite{zhang2018adversarial}, CatAAE \cite{liu2018unsupervised}, DSN \cite{sun2018sparse} and ADASYN \cite{lee2017application}. To implement these methods we followed their structure\footnote{\textcolor{black}{https://github.com/openai/improved-gan}}, and for the false detection rate between normal $\&$ faulty samples we defined the ratio $\frac{1-D_g}{D_g}$. For a fair comparison, we trained the baselines and our proposed model with the same system setting. Our experiments are carried out on several relative datasets. We evaluated our proposed model over two cases; the first case includes binary imbalanced class, while the second case includes a multiclass imbalance problem. More details are given in the next section. Recall that, in the training phase, we use our proposed GAN to generate minority class samples from mixture distribution and in the testing phase we show that our designed discriminator is able to correctly distinguish the faulty samples from the normal ones.
\begin{table*}
\centering
\caption{Comparative results on IMS dataset. {\em Mi} refers to Minority samples and {\em Mj} refers to majority samples. Best results are bold \protect \footnotemark}
\begin{tabular}{p{.15\textwidth}|p{.08\textwidth}p{.08\textwidth}p{.05\textwidth}p{.05\textwidth}p{.05\textwidth}| p{.08\textwidth}p{.08\textwidth}p{.05\textwidth}p{0.05\textwidth}p{.05\textwidth}}
\hline 
& \multicolumn{5}{c}{S1} & \multicolumn{5}{c}{S2} \\   \cline{1-10}
\hline \hline
Methods & Mi testing &  Mj testing  & Training set  & Testing set& $G_{mean}$ & Mi testing & Mj testing & Training set & Testing set& $ G_{mean}$  \\ [0.5ex] 
\hline
SVM$_{RBF}$ & \textcolor{blue}{0.5248} & 0.9707&	0.7385&0.7631&\textcolor{ForestGreen}{0.9453}&\textcolor{blue}{0.6381}&0.9826&0.9139&0.9233&\textcolor{ForestGreen}{0.8663}   \\  [0.5ex]
DNCNN \cite{jia2018deep} &  \textcolor{blue}{0.8693}&1.000&0.8991&	0.9287&\textcolor{ForestGreen}{0.9183}&\textcolor{blue}{0.7529}&1.000&0.9364&0.918&\textcolor{ForestGreen}{0.8997} \\
Jiang et al.\cite{jiang2019novel} & \textcolor{blue}{0.8817}&1.000&0.923&0.9128&\textcolor{ForestGreen}{0.9321}&\textcolor{blue}{0.8916}&1.000&0.9752&0.9623&\textcolor{ForestGreen}{0.9141} \\
Shen et al. \cite{shen2013fault} & ---& ---& 0.9813 & 0.9427 & \textcolor{ForestGreen}{0.9607} & --- & --- & 0.9985 & 0.9341 & \textcolor{ForestGreen}{0.946} \\
Chen et al.\cite{chen2020deep} &\textcolor{blue}{0.8336}&0.9973&0.8966&0.9027&\textcolor{ForestGreen}{0.9399}&\textcolor{blue}{0.8884}&0.9994&0.9418&0.9532&\textcolor{ForestGreen}{0.8993} \\
MDIAN \cite{wang2020multi} & --- & --- & 1.000 & 0.946 & \textcolor{ForestGreen}{0.9347} & --- & --- & 0.9938 & 0.9826 & \textcolor{ForestGreen}{0.9018} \\
NSVDD \cite{liu2020semi}&\textcolor{blue}{0.8505}&1.000&0.9726&0.9348&\textcolor{ForestGreen}{0.9594}&\textcolor{blue}{0.8737}&0.9917&0.9905&0.9711&\textcolor{ForestGreen}{0.924}  \\
{\bf MoGAN} &\textcolor{blue}{\bf 0.9123}&	1.000&0.9853&0.9539&\textcolor{ForestGreen}{\bf 0.9789}&\textcolor{blue}{\bf 0.9431}&0.9857&0.9986&0.9838&\textcolor{ForestGreen}{\bf 0.9601} \\
\hline
\end{tabular}
\label{tab:3}
\end{table*} 
\footnotetext{The results of \cite{wang2020multi}, \cite{shen2013fault} and \cite{chen2020deep} are taken from their works.}

\subsection{Dataset preparation and implementation setting}
For developing an efficient model for the fault diagnosis, providing good data is necessary. In this paper, the
datasets are selected based on the given information in \cite{zhang2019machine}.
\begin{itemize}
\item {\bf CWRU dataset \footnote{\textcolor{black}{http://csegroups.case.edu/bearingdatacenter/home}}}, Case Western Reserve Lab is simulated under three working conditions: (load $1= 1hp/1772rpm$; load $2=2hp/1750rpm$; load $3=3hp/1730rpm$). In each working condition, four status are collected, normal, inner fault (IF), outer fault (OF), and ball fault. We randomly selected 2048 samples after removing coefficients. We grouped these samples based on the mentioned conditions as A, B, C respectively. Each condition contains ten classes (one normal + nine faulty samples), with 800 samples in each. In fact, each dataset contains 8000 samples including normal and faulty. More details are given in Table \ref{tab:1}.
\item {\bf Case 1 \footnote{\textcolor{black}{http://www.industrial-bigdata.com}}, Wind turbine freezing failure forecast (WTFF)}, held by MIIT of China, it contains $28-dimensional$ time
series. The data are binary labeled, as normal and faulty samples. The number of normal samples is “25768”, while the faulty samples are around “1554”, wherein the imbalancedness ratio can be about 1:17. For the testing set, we extract all types of data (contains both the normal $\&$ fault) from the 90,000th to 149,999th timestamp, and the rest are used for training.
\item {\bf Case 2 \footnote{\textcolor{black}{https://www.phmsociety.org/}} Industrial plant failure detection (IPF)}, is provided by PHM society in 2015. The data contains eight dimensional time series and is a multiclass data with six type of fault. The highest imbalancedness ratio may reach to 1:1000. Table \ref{tab:2}, shows more details about used datasets.
\item{\bf IMS bearing dataset \footnote{\textcolor{black}{https://ti.arc.nasa.gov/tech/}}}, is a well-known data which is provided by the Center for Intelligent Maintenance Systems (IMS), University of Cincinnati, and generated from the “Prognostics Data Repository of NASA”. Two different test-to-failure experiments are conducted as to achieve following datasets: outer race fault ($S1$ data) and ball fault ($S2$ data). We follow \cite{mao2017online} \cite{liu2020semi}, and use their collected data in this experiments. $S1$ contains 6000 samples in which we used 4300 as training and the rest as testing. In the training set, there are 3800 normal samples, 500 faulty samples. We chose the imbalance ratio around $10:1$. The same process also is executed to generate the ball fault dataset $S2$.   
\end{itemize}
It is worth to mention that, the implementations of the network are conducted on a workstation equipped with an
Intel i7-6850K CPU with a 64 GB Ram and an {\tt NVIDIA GTX} Geforce 1080 Ti GPU and the operating system is Ubuntu 16.04.

\begin{table}
\centering
\caption{Precision, Recall, FAM - Different Resampling Methods using Case 1.}
\begin{tabular}{p{.4\textwidth}p{.1\textwidth}p{.12\textwidth}p{.1\textwidth}}
\hline
Methods & Recall & Precision & FAM \\
\hline \hline 
Original GAN \cite{goodfellow2014generative}& 0.4973 & 0.6873 & 0.7297 \\
MixtureGAN (ours) & {\bf 0.8902} & {\bf 0.9427} &{\bf  0.9064} \\
CaAE \cite{ren2020novel} &  0.8641 & 0.8873 & 0.8619  \\
RJAAN \cite{jiao2020residual}& 0.8814 & 0.9106 & 0.8653 \\
B-SMOTE \cite{han2005borderline}& 0.8452 & 0.6470 & 0.6941 \\
EMICIL \cite{razavi2017integrated} & 0.7723 & {\textcolor{blue}{0.9275}} & 0.8799 \\
EWMOTE \cite{lin2018adaptive} & 0.7691 & 0.9138 & 0.8827 \\
WSMOTE \cite{zhang2016transfer} & 0.8502 & {\textcolor{blue}{0.9336}} & {\textcolor{blue}{0.8993}} \\ 
\hline
\end{tabular}
\label{tab:4}
\end{table} 
\begin{table}
\centering
\caption{Precision, Recall, FAM - Different Resampling Methods using Case 2.}
\begin{tabular}{p{.4\textwidth}p{.1\textwidth}p{.12\textwidth}p{.1\textwidth}}
\hline
Methods & Recall & Precision & FAM \\
\hline \hline 
Original GAN \cite{goodfellow2014generative}&0.6733 & 0.7859 & 0.6983 \\
MixtureGAN (ours)& {\bf 0.8531} & {\bf 0.8946} & {\bf 0.8730}\\
CaAE \cite{ren2020novel} & 0.8016 & 0.8236 & 0.8253 \\
RJAAN \cite{jiao2020residual} & 0.8375 & 0.8514 & 0.8432 \\
B-SMOTE \cite{han2005borderline}&0.8531 & 0.8946 &0.8730 \\
EMICIL \cite{razavi2017integrated} & 0.8305 & 0.8426& 0.8342 \\
EWMOTE \cite{lin2018adaptive} & {\textcolor{blue}{0.8609}} & 0.8527 & {\textcolor{blue}{0.8667}} \\
WSMOTE \cite{zhang2016transfer} & 0.8436 & 0.8709 & 0.8561 \\ 
\hline
\end{tabular}
\label{tab:5}
\end{table} 

\subsection{Assessment Metrics}
\label{subsec:1}
For imbalance class problems, overall accuracy and mean square error are not appropriate. Instead, we used balanced accuracy which is defined in \cite{zhang2016transfer} as: 
\begin{equation}
\begin{aligned}
BAC& = \frac{1}{2}(sensitivity + specificity) &&\\
G_{mean}&= \sqrt{sensitivity \times specificity} && \\
f_{measure}& = (1+\alpha^2)\frac{sensitivity\times specificity}{\alpha \times sensitivity+ specificity} &&
\end{aligned}
\end{equation}
where $\alpha$ is generally taken to be 1) in order to provide a comprehensive evaluation. In the confusion matrix, the left column represents minority class samples of the original dataset and the right column is the majority one. The proportion of these columns will be the class distribution in each dataset. The metrics have been calculated as follows.

\section{Results and Analysis}
\label{sec:6}
Motivated by GAN models, we present a novel architecture to generate minority class samples from a mixture of data distribution in order to restore a balance in imbalanced datasets. We also used a feature matching loss function to handle such mixture data distribution. Figure \ref{fig:3}, shows the confusion matrix for GAN based Feature matching loss function on the WTFF dataset. From left to right, these values represent true positive rate, false-positive rate, false-negative rate, true negative rate. These values convey that, the normal samples give more than 95\% TPR, and the faulty samples around 89\%. It means, our proposed mixture data distribution can produce completely different faulty samples which are recognizable from the normal sets.

\begin{figure*}[h]
    \centering
   \begin{minipage}{0.42\textwidth}
\centering
\includegraphics[width=1\textwidth]{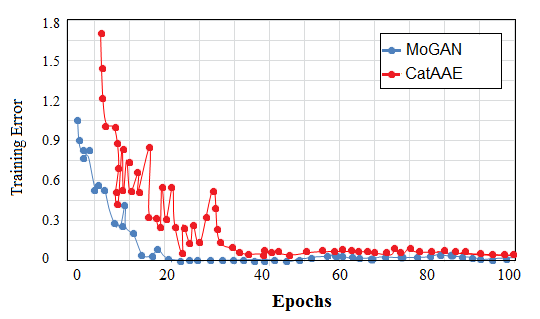}
\end{minipage} 
  \begin{minipage}{0.4\textwidth}
        \centering
\includegraphics[width=1\textwidth]{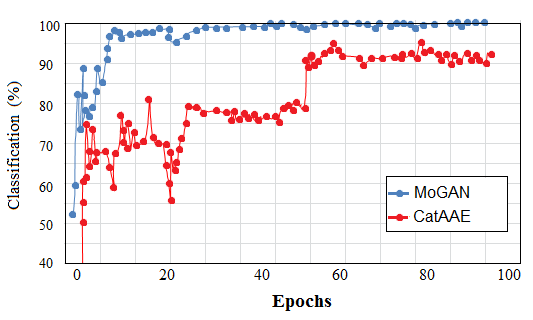}
\end{minipage}
       \caption{Performance of training error and classification rate of MoGAN and CatAAE \cite{liu2018unsupervised} on the Case 1.}
    \label{fig:6}
\end{figure*}

\begin{figure*}
    \centering
\includegraphics[width=15cm]{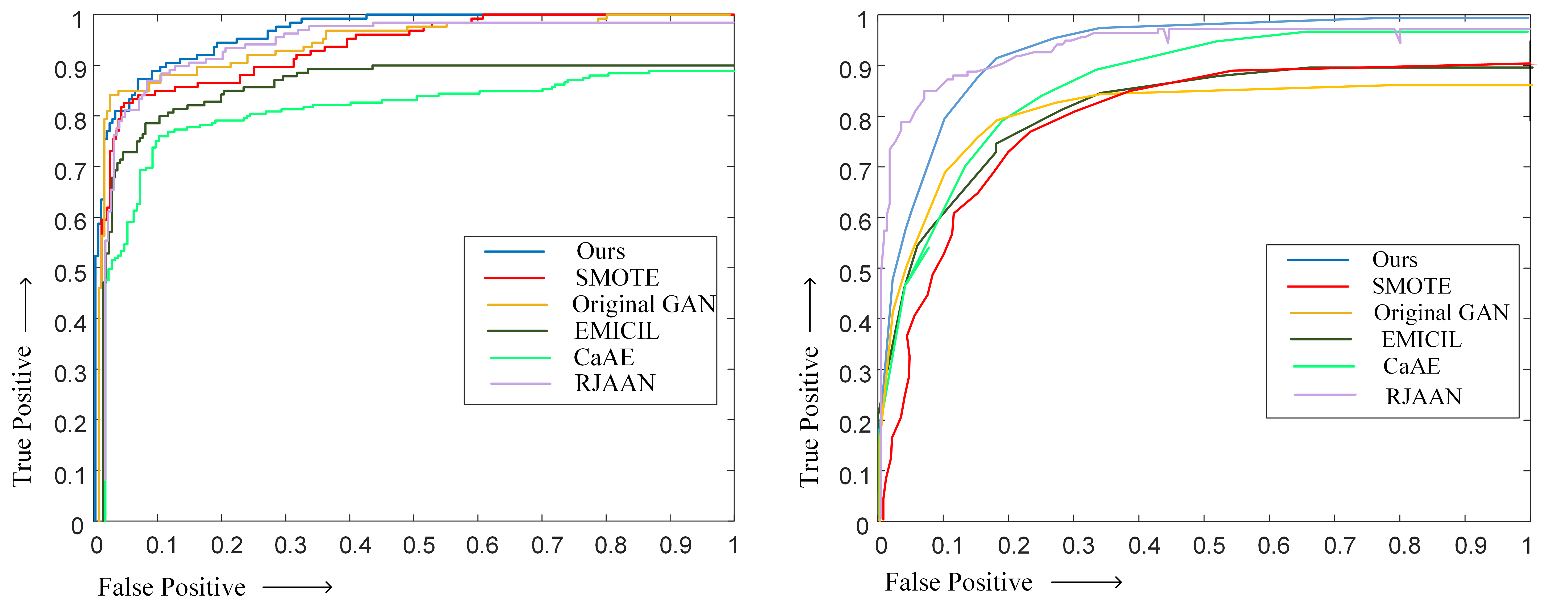} 
    \caption{The ROC curve comparisons between various methods using Case 1 and Case 2. We used SMOTE  \cite{han2005borderline}, GAN \cite{goodfellow2014generative}, EMICIL \cite{razavi2017integrated}, CaAE \cite{ren2020novel}, and RJAAN \cite{jiao2020residual} as baselines.}
    \label{fig:5}
\end{figure*}

\begin{figure*}
\vspace{-15mm}
    \centering
    \includegraphics[width=18.5cm]{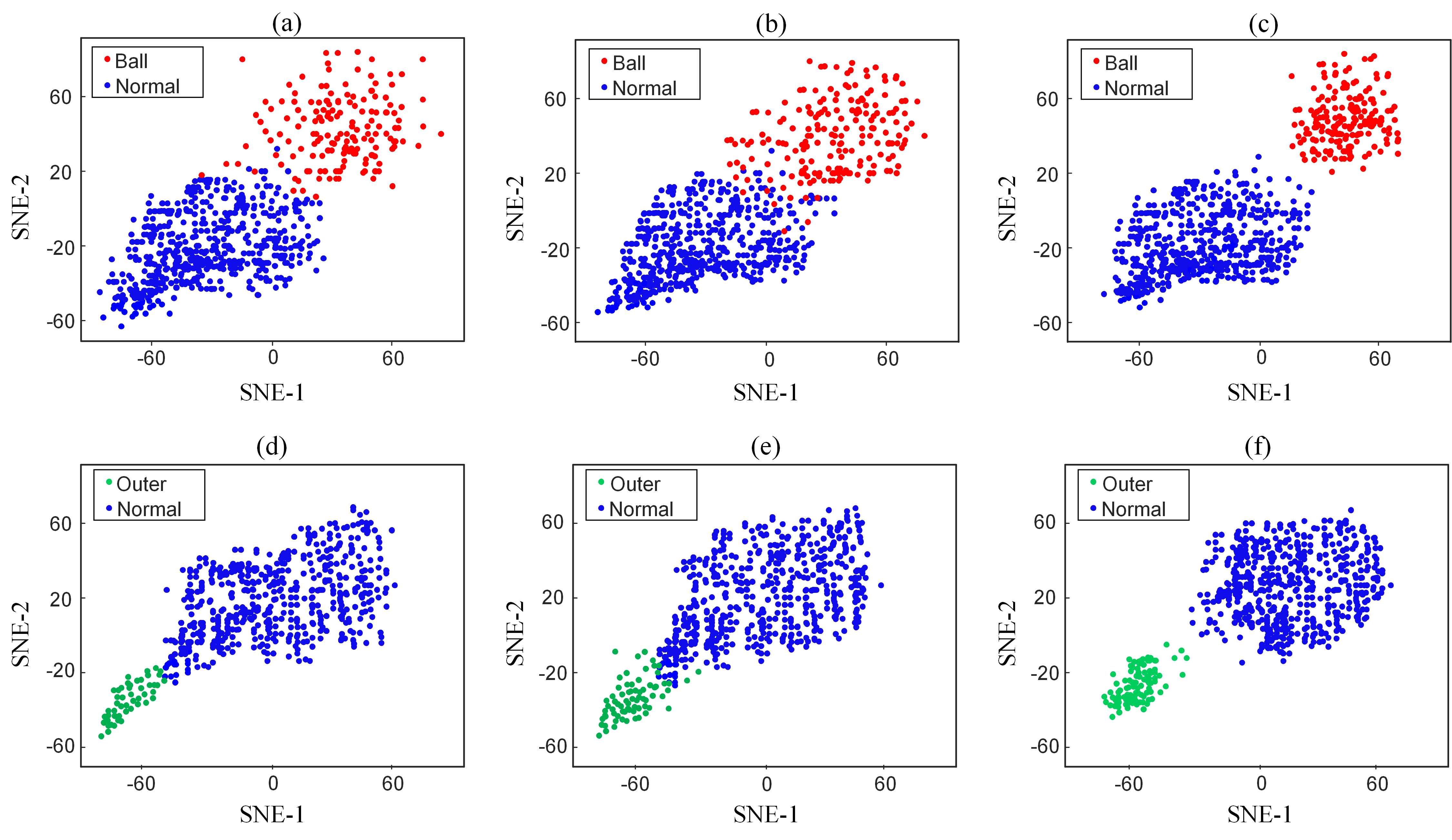} 
    \caption{The t-SNE feature visualization. (a, d) A2CNN \cite{zhang2018adversarial}, (b, e) RJAAN \cite{jiao2020residual} ; (c, f) MoGAN.}
    \label{fig:14}
\end{figure*}

\begin{figure*}
\vspace{-20mm}
    \centering
    \includegraphics[width=18cm]{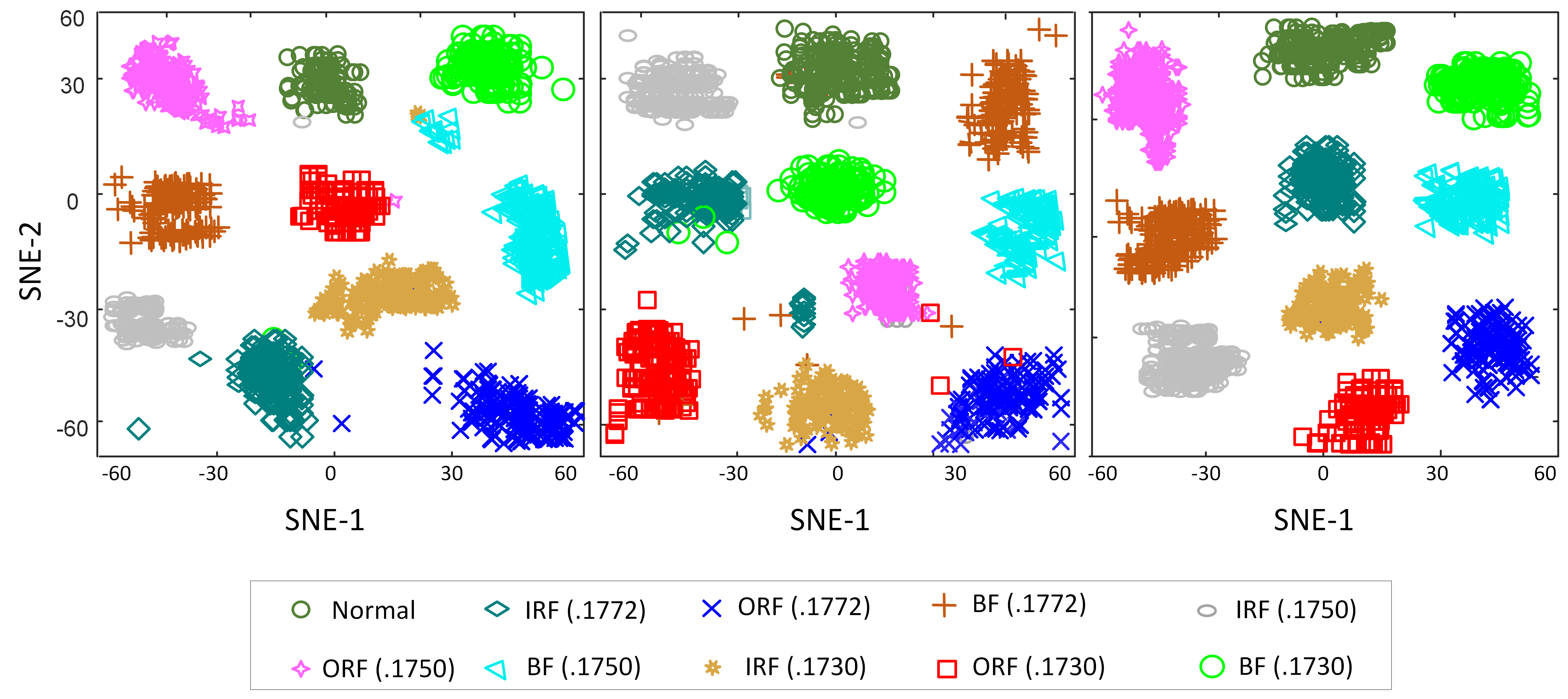} 
    \caption{The t-SNE feature visualization. (a) A2CNN \cite{zhang2018adversarial}, (b) RJAAN \cite{jiao2020residual} ; (c) MoGAN.}
    \label{fig:15}
\end{figure*}

\begin{figure*}[!t]
    \centering
    \includegraphics[width=16cm]{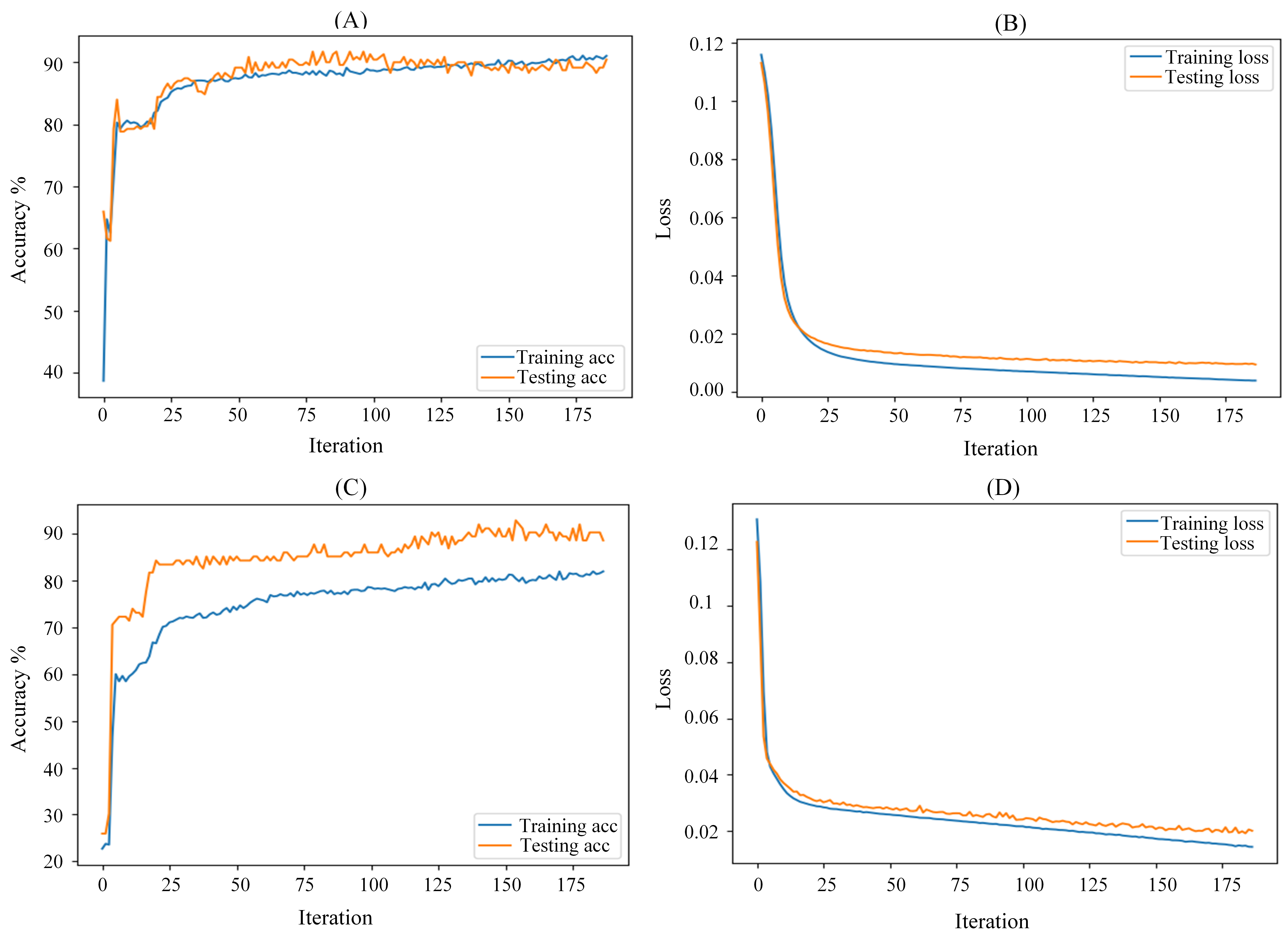} 
    \caption{Relationship between accuracy and loss while applying cross-training on bearing 1$\leftrightarrow2$ datasets. (A) and (B) bearing $1\rightarrow2$, (C) and (D) bearing $2\rightarrow1$.}
    \label{fig:16}
\end{figure*}

In addition, we conducted several experiments in which the results are shown in Figures \ref{fig:3} - \ref{fig:16} . We also compared the proposed generative model with different resampling methods and the results are summarized in Table \ref{tab:3} - \ref{tab:5}. \\

{\bf SVM performance on the simulated dataset}:  The support vector machine is a statistical classifier introduced by \cite{vapnik1999overview}. In this paper, we show the performance of applying SVM to the generated dataset. In our work, the linear kernel is not appropriate to separate faulty samples
from normal ones. Therefore, we also use a radial basis function to properly separate the labeled datasets.

Briefly, {\tt RBF kernel} is calculated as $E_{ij}=\exp (-\frac{\|\alpha\|^2}{2 \rho^2})$, where, $\rho$ is a user-defined parameters and $\|\alpha\|$ is the Eclidean distance between two samples $x_i$ and $x_j$. We followed \cite{zareapoor2018kernelized} to implement the hyperparameters of RBF on the WTFF dataset. Figure \ref{fig:3} (right), shows the confusion matrix for the labeled dataset (balanced dataset) using SVM based RBF kernel as a classifier. Although the true positive accuracy rate is around70\%, the TNR is less than 55\%. We can
conclude that the normal samples with 66\% TPR are well recognized, but the faulty due to 62\% TPR, they couldn’t well identify. SVM despite using a balanced dataset gives a poor performance for TPR. Figure \ref{fig:3} (left) shows better classification performance compared to SVM.\\

\begin{figure*}
\vspace{-5mm}
    \centering
   \includegraphics[width=18cm]{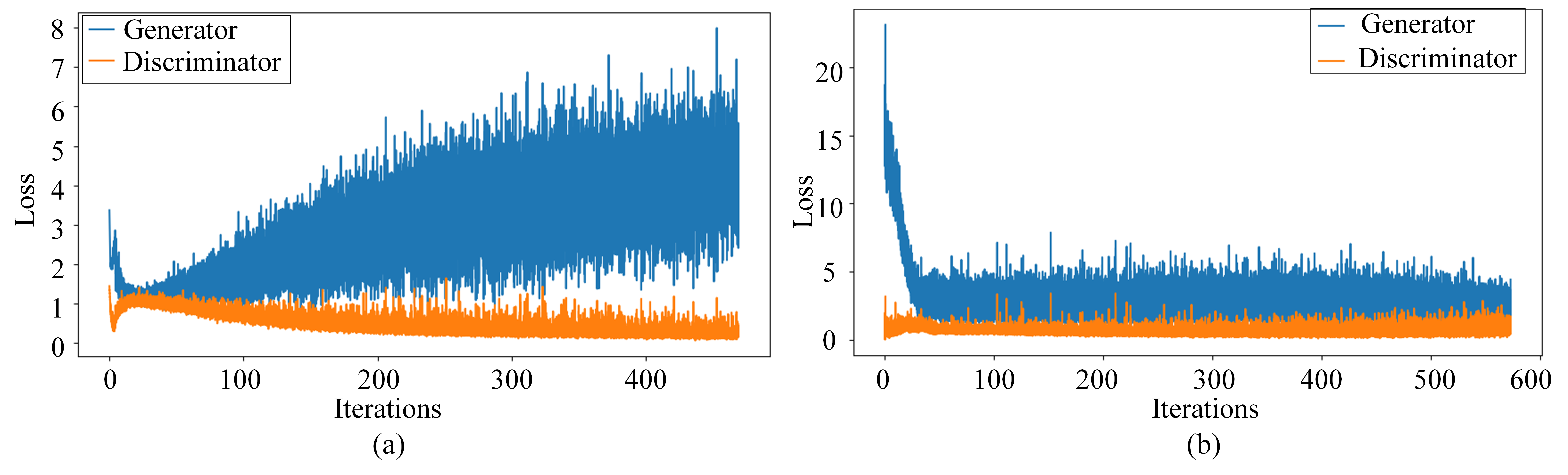}
    \caption{The convergence analysis of MoGAN. Generator and Discriminator loss during training. (Left): Case 1 and (Right): Case 2}
    \label{fig:12}
\end{figure*}

{\bf Experiments on IMS bearing data}: In this section we show our models' functionality using IMS dataset. We train our model over 70\% of available data to evaluate the prediction rate of our model. Table \ref{tab:3} summarizes the prediction accuracy in each experiment. For fair comparisons, we used several states of the arts methods, and compare our model with them. On order to keep the results robust, we used the average of 50 trials as the final result. It is worth to mentioning that for the highly imbalanced data, minority sample prediction is more importance than the majority. As we can see in Table \ref{tab:3}, the minority testing accuracy of our model is around 4\% and 6\% higher than the second best results in each datasets. $G_{mean}$ as defined in section \ref{subsec:1} is a promising evaluation metrics for the classification of imbalanced dataset. The results convey that, G-mean value of MoGAN outperforms the other baselines. In addition to these results we also used different imbalanced ratio rate such as: $5:1  \&  3:1$, and compared our results with the same baselines. Due to reducing the imbalancedness ratio in the training, we achieve $0.983$ score in $G_{mean}$, and the results of NSVDD \cite{liu2020semi} and DNCNN \cite{jia2018deep} are close to our results as: $0.973$ and $0.977$ respectively. \\

In Figure \ref{fig:4}, we plot the balanced accuracy curves over S1 and S2 dataset based on the number of hidden nodes.  We compares our proposed model with Five baselines that exploited the deep neural network in their architectures. The results are achieved from the average 10 trials. As it observes, by growthing the number of hidden nodes, our proposed model has a smooth and robust performance.  In both datasets, our model outperforms the other baselines in terms of the "minority testing accuracy". \\

{\bf Diagnosis results of different methods}: We conducted several experiments for fault diagnosis problem. Our proposed model is trained and tested using two different settings. First, we use Border-SMOTE \cite{han2005borderline} (B-SMOTE) and other resampling techniques to traditionally resample the minority class instances and simultaneously applied our discriminator to find the represented class for each instance. Second, we use our proposed mixture data distribution (including minority and majority class samples) to generate minority class samples and then our proposed MoGAN is trained on the new dataset. As suggested in \cite{lin2018adaptive}, we used Recall, Precision, and FAM (average of AUC, MCC, and F-measure) as evaluated metrics. The results are given in Table \ref{tab:4}.\\

Although, using SMOTE displays better TPR for the normal samples, but it performs very poor for the faulty samples due to the less number of samples. From the results, it observes that there is a significant performance on both the normal and faulty samples by using resampling techniques. SMOTE performs a bit better than GAN, EWMOTE, and EMICIL on recall. It may due to generating wrong faulty samples that fall insie the normal class region. Therefore, this can improve the recall, accordingly reduce the precision and FAM, because some of the faulty samples fall inside of the majority region. The second best results can be found for WSMOTE.
\begin{figure*}
    \centering
\includegraphics[width=17cm]{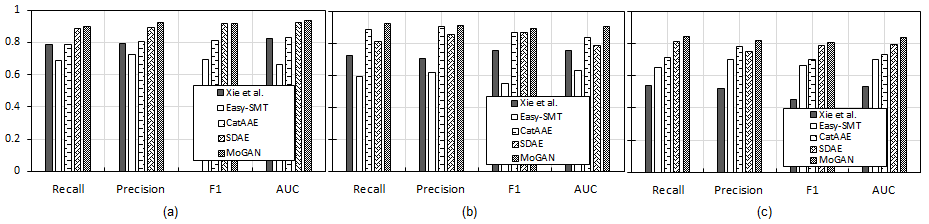} 
    \caption{The comparison results between various fault diagnosis techniques based on Recall, Precision, F1, and AUC. The baselines are; Xie et
al.\cite{xie2018imbalanced}, Easy-SMT \cite{wu2018integrated}, SDAE \cite{lu2017fault}, CatAAE \cite{liu2018unsupervised}; left to right-CWRU dataset (Data A, B and C).}
    \label{fig:7}
\end{figure*}
\begin{figure*}
    \centering
\includegraphics[width=17cm]{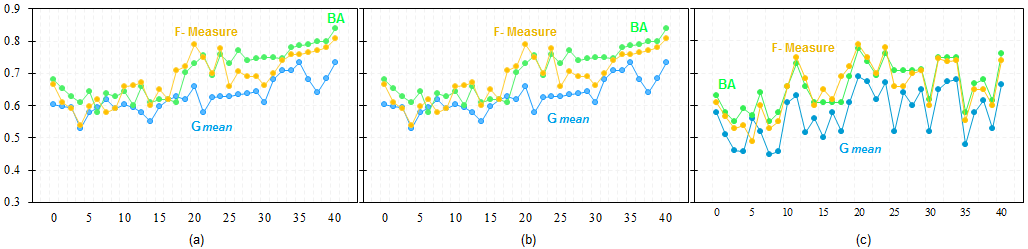} 
    \caption{$BA$ (Balanced accuracy) values, $F_{measure}$ and $G_{mean}$ values in different epochs on CWRU dataset using MoGAN; left to right: Data A, B, and C. We note that, the iteation is up to "$\times10^2$".}
    \label{fig:8}
\end{figure*}

Based on all evaluation metrics, generated samples using our model are closer to the original samples comparing to other resampling techniques. We also use {\tt Case 2} - which is the multiclass dataset - to compare the various oversampling methods for imbalanced learning. The results summarized in Table \ref{tab:5}. In contrast to Table \ref{tab:4}, in Case-2 , SMOTE gives a poor performance on recall, which means that it generated noisy and useless minority samples that leads to misclassification problem, thus decrease the recall.

However, the recall of original GAN, WSMOTE, and EMICIL is much better than SMOTE, but worse in precision and FAM scores. It is worth to mention that, WSMOTE is better than EMICIL and EWSMOTE on precision, which means the decision boundary, is prone to the majority class region in order to generate the new samples. As it observes, the proposed model performs well on imbalanced multiclass learning in all the assessment measures. We also plot the ROC graph in Figure \ref{fig:5}. We used {\tt Case 1} for binary classification and {\bf Case 2} for the multiclass classification. We used SMOTE  \cite{han2005borderline}, GAN \cite{goodfellow2014generative}, EMICIL \cite{razavi2017integrated}, CaAE \cite{ren2020novel}, and RJAAN \cite{jiao2020residual} as baselines. The results show, our proposed model performs better than other sampling and none-sampling techniques for both the binary and multiclass examples. However, in the binary classification, the second-best result is for the Original GAN, and SMOTE. While in the multiclass dataset, the second-best result gives to RJAAN which has a comparable performance with MoGAN.

In addition, we compare our proposed MoGAN with CatAAE, since they have a similar setting in sense of using GAN to generate minority samples and use discriminator to distinguish the faulty samples from the normal samples. The results are given in Figure \ref{fig:6}. We achieved the results on the Case 1 dataset. It observed that the proposed model has a less training error compares to CatAAE, also it reaches the convergence after a few epochs (faster convergence).

{\bf Feature visualization}: In this expriment we demonstrate the effectiveness of our model, and illustrate its versatility by tackling imbalancdeness problem, using t-SNE \cite{maaten2008visualizing} to visualize the feature representations. We used RJAAN \cite{jiao2020residual} and A2CNN \cite{zhang2018adversarial} as baseline, and compared our proposed MoGAN with these models. The expriment is evaluated on IMS and CWRU datasets. The results is presented in Figure \ref{fig:14} and Figure \ref{fig:15}. From the results can be observed that, MoGAN and RJAAN can extract more effective features of each health status which further help to improve the accuracy performance. However, our proposed MoGAN method can better aggregate features of the same health condition for accurate classification.The distribution of features extracted by the MoGAN is better than other models. The second best results are obtained by RJAAN model. 

In Figure \ref{fig:16} we show the bearing degradation assessment by using IMS (bearing 1 and 2) dataset on healthy indicators while using cross-training and testing. We have shown the training-testing accuracy and loss while training on one dataset and testing on the other one. Figure \ref{fig:16} demonstrates the relation between accuracy and loss during the cross-training of the two bearings of IMS datasets ($1 \Leftrightarrow2$). The results indicate that applying cross-training helps to minimize the loss and also increases the adaptation accuracy. In this evaluation, the best results are achieved from $1\Leftrightarrow2$, since, the number of the training samples in the bearing $1$ dataset is more than bearing $2$ dataset and the testing samples in bearing $2$, have low complexity.

Figure \ref{fig:12} represents the stability of the proposed model. The figure contains two subplots, the discriminator loss (orange) and the generator loss (blue). It can be observed, both losses are erratic before around iteration 50.  As we used pre-trained generator in our model, it prevents faster convergence of the discriminator compare to the generator at the early epoch and maintaining the stabilized learning process of GANs and reducing mode collapses.

Next, we show the proposed model effectiveness on the CWRU datasets. We compared our model with different techniques based on fault diagnosis. We evaluated the performance of our model with other baselines on several assessment metrics, which are defined in section \ref{subsec:1} like, BA, Recall, Precision, AUC, G-mean, and F-measure. The results are shown in Figure \ref{fig:7} and Figure \ref{fig:8}. For both evaluations, we used CWRU with three data categories as A, B, and C.

In dataset A, our model performs better than baselines in all evaluation metrics, and the second-best goes for CatAAE and SDAE on precision. The evaluation scores convey that, these methods correctly classified the faulty and normal samples, but they failed to generate dense and real-look faulty samples to improve the recall and AUC. For the dataset B and C, our model has a compatible performance with SDAE. However, SDAE in both the datasets failed to generate faulty samples that fall in the minority region. Easy-SMT has the worst performance on
all three datasets. The work by Xie at al.\cite{xie2018imbalanced} has a compatible performance with other baselines only on the dataset A, for the other datasets, it doesn’t show any progress. For more details please see Figure \ref{fig:7}, \ref{fig:8}. 
\begin{figure*}[!t]
    \centering
    \includegraphics[width=15cm]{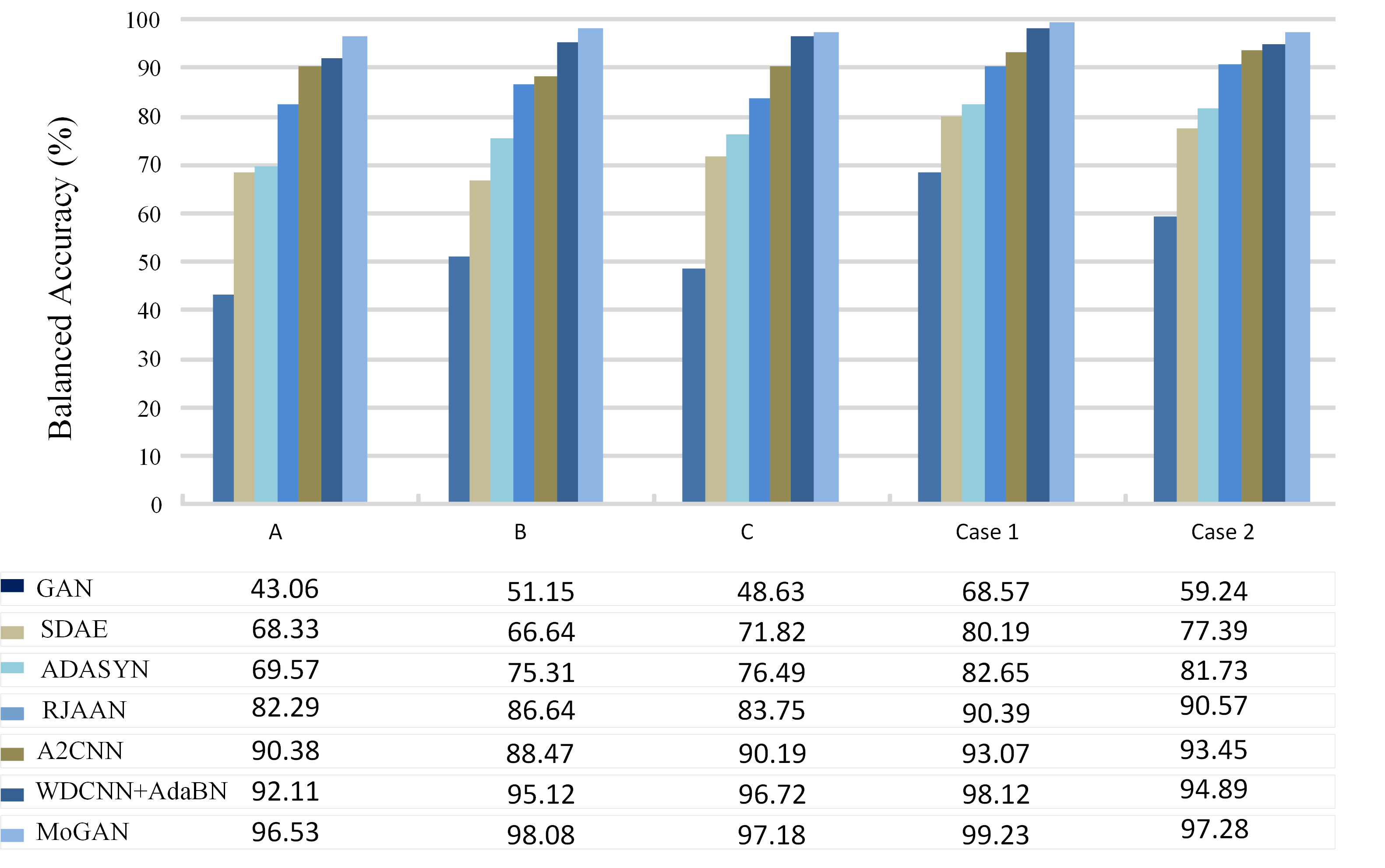} 
    \caption{BA: Balanced Accuracy (\%) on different datasets. We used GAN \cite{goodfellow2014generative}, SDAE \cite{lu2017fault}, ADASYN \cite{lee2017application}, RJAAN \cite{jiao2020residual}, A2CNN \cite{zhang2018adversarial} and WDCNN+AdaBN \cite{zhang2017new} as baselines.}
    \label{fig:9}
\end{figure*}
\begin{figure*}[!htb]
    \centering
    \includegraphics[width=19cm]{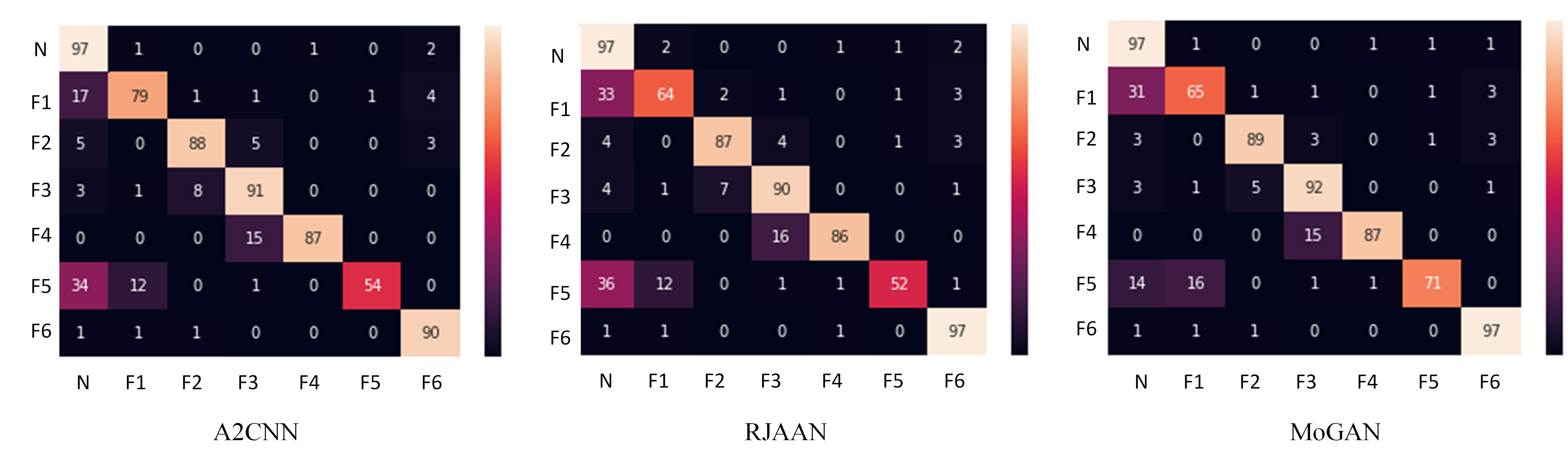} 
    \caption{Confusion matrix of A2CNN \cite{zhang2018adversarial}, RJAAN \cite{liu2018unsupervised}; and MoGAN.}
    \label{fig:10}
\end{figure*}

We also illustrate the performance of our proposed model base on mixture data distribution over several synthetic sampling techniques to identify the faulty samples, in Figure \ref{fig:9}. We compared our model with several relative baselines like; SDAE \cite{lu2017fault}, ADASYN \cite{lee2017application}, RJAAN \cite{jiao2020residual}, A2CNN \cite{zhang2018adversarial}, WDCNN+AdaBN \cite{zhang2017new} and original GAN \cite{goodfellow2014generative}. We can find that for Case 1 and dataset C, the performances of our model and RJAAN are very close. However, for Case 2, Dataset A and B, our model gives better scores. As the results show, simple GAN performs poorly in all datasets with average accuracy being around (53\% - 62\%). 
It proves that simple GAN is not suitable for fault diagnosis techniques, in particular, multiclass classification problems. For binary classification, GAN obtained 68\% BA which is smoothly better than other scores. The second worse gives to SDAE, which has poor performance for multiclass datasets, but for the binary (Case 1) shows a good result. WDCNN with using AdaBN achieved 92\% balanced accuracy which is obviously greater than other baselines. This result conveys that the samples learned by WDCNN+AdaBN are more invariant than the samples
learned by the other baselines. In addition, by comparing our model with A2CNN we can find that in each dataset, the performance of our model is superior to A2CNN. This means that generating minority samples from the mixture data distribution and using feature matching function in order to handle the mixture distributions, can significantly improve the fault diagnosis under different datasets. It is also interesting that for binary datasets such as case 1, the fault diagnosis balanced accuracy of the WDCNN+AdaBN, RJAAN, and A2CNN is a bit better than others.

Finally, we show the classification result of each class in Case 2 datasets including normal and six faulty samples (the dataset details are given in Table \ref{tab:2}) in Figure \ref{fig:10}. $F1 - F6$ indicates the number of faults and $N$ is the normal sample. $F6$ is categorized as the low imbalanced ratio ($ratio rate around 1:3$), $F1, F2$, and $F3$ are categorized as
medium imbalance ratio (around  1: 20) and $F4, F5$ is categorized as the severe imbalanced ratio (around 1: 1000). From the results it can be easily inferred that, the proposed MoGAN performs better for low and medium imbalanced ratio, but a little get worse for severe imbalancedness (but still it is better than the baselines). Moreover, our proposed model is more robust compares to the other baselines on recognizing the faulty samples. In particular, it shows a good performance with a high imbalanced ratio compares to the RJAAN and A2CNN.

\section{Conclusion}
\label{sec:7}
The ability to recognize the faulty samples in an imbalanced dataset is very important for many classification systems. The proposed MoGAN is an effective end to end model for {\em simultaneous classification and fault detection} in imbalanced fault diagnosis. To restore balance in the imbalanced dataset we used {\em mixture data distribution} to generate minority class samples. In the case of fault diagnosis with a mixture of data distribution, we proved that the discriminator can be considered as a {\em fault detector}. We also trained our generator which contains different data distribution with feature matching loss function. In this way, the generator ables to generate samples in low-density areas of data and helps to the discriminator for better observation of detection. Three sets of data are used to evaluate our proposed model. The comprehensive experiments show, the proposed solution is comparable to other popular
fault detection techniques, even in some cases outperforms them. We also trained a SVM classifier on the generated samples for classification task. The results report that SVM has poor performance comparing to our proposed model. However, the obtained result from the generated data (66\% - 51\%) is better than the result of using the original dataset (48\% - 27\%). This clearly reports that our proposed generator with mixture data distribution is able to generate informative samples that are useful for the classification. 

For the future, this work will be extended in order to reduce the dimensionality to could improve fault diagnosis approaches under different conditions. We would like to develop a new loss function to handle mixture data distribution that enriches the generator performance and accordingly improve the discriminator’s detection rate.

\section*{Acknowledgment}
This research is partly supported by NSFC, China (No: 61806125, 61977046), Committee of Science and Technology, Shanghai, China  (No. 19510711200).

\bibliography{manuscript}

\end{document}